\documentclass[lettersize,journal]{IEEEtran}
\usepackage{graphicx}
\usepackage{amsmath,amsfonts}
\usepackage{array}
\usepackage[caption=false,font=normalsize,labelfont=sf,textfont=sf]{subfig}
\usepackage{textcomp}
\usepackage{stfloats}
\usepackage{url}
\usepackage{verbatim}

\usepackage{bm}
\usepackage{biblatex} 
\usepackage{color}
\usepackage{hyperref}
\usepackage{algorithm}%
\usepackage{algorithmicx}%
\usepackage{algpseudocode}%
\usepackage{bbm}
\usepackage{cuted}
\usepackage{booktabs}%
\usepackage{xcolor} 
\usepackage{colortbl} 

\usepackage{multirow}
\usepackage{romannum}

\def\BibTeX{{\rm B\kern-.05em{\sc i\kern-.025em b}\kern-.08em
    T\kern-.1667em\lower.7ex\hbox{E}\kern-.125emX}}
\usepackage{balance}

\usepackage{parskip}
\begin{document}

\title{Brain-inspired analogical mixture prototypes for few-shot class-incremental learning}

\author{Wanyi Li, \textit{Member, IEEE}, Wei Wei, \textit{Member, IEEE}, Yongkang Luo, \textit{Member, IEEE}, Peng Wang, \textit{Member, IEEE}
\thanks{Manuscript received Month, Day, 2025; This work was supported by the National Natural Science Foundation of China (grants 61771471, 91748131, and 62006229) and the Strategic Priority Research Program of the Chinese Academy of Sciences (grant XDB32050106).(Corresponding author: Peng Wang).

Wanyi Li is with State Key Laboratory of Multimodal Artificial Intelligence Systems, Institute of Automation, Chinese Academy of Sciences, Beijing 100190, China (e-mail:wanyi.li@ia.ac.cn).

Wei Wei is with State Key Laboratory of Multimodal Artificial Intelligence Systems, Institute of Automation, Chinese Academy of Sciences, Beijing 100190, China, and also with the School of Artificial Intelligence, University of Chinese Academy of Sciences, Beijing 100049, China (e-mail: wei.wei2018@ia.ac.cn)

Yongkang Luo is with State Key Laboratory of Multimodal Artificial Intelligence Systems, Institute of Automation, Chinese Academy of Sciences, Beijing 100190, China (e-mail:yongkang.luo@ia.ac.cn).

Peng Wang is with State Key Laboratory of Multimodal Artificial Intelligence Systems, Institute of Automation, Chinese Academy of Sciences, Beijing 100190, China, with the CAS Center for Excellence in Brain Science and Intelligence Technology, Chinese Academy of Sciences, Shanghai 200031, China, and also with the Centre for Artificial Intelligence and Robotics, Hong Kong Institute of Science and Innovation, Chinese Academy of Sciences, Hong Kong 999077, Hong Kong (e-mail: peng\underline{~}wang@ia.ac.cn).}}

\markboth{Journal of \LaTeX\ Class Files,~Vol.~18, No.~9, September~2020}%
{How to Use the IEEEtran \LaTeX \ Templates}

\maketitle
\begin{abstract}
Few-shot class-incremental learning (FSCIL) poses significant challenges for artificial neural networks due to the need to efficiently learn from limited data while retaining knowledge of previously learned tasks. Inspired by the brain's mechanisms for categorization and analogical learning, we propose a novel approach called Brain-inspired Analogical Mixture Prototypes (BAMP). BAMP has three components: mixed prototypical feature learning, statistical analogy, and soft voting. Starting from a pre-trained Vision Transformer (ViT), mixed prototypical feature learning represents each class using a mixture of prototypes and fine-tunes these representations during the base session. The statistical analogy calibrates the mean and covariance matrix of prototypes for new classes according to similarity to the base classes, and computes classification score with Mahalanobis distance. Soft voting combines both merits of statistical analogy and an off-shelf FSCIL method. Our experiments on benchmark datasets demonstrate that BAMP outperforms state-of-the-art on both traditional big start FSCIL setting and challenging small start FSCIL setting. The study suggests that brain-inspired  analogical mixture prototypes  can alleviate catastrophic forgetting and over-fitting problems in FSCIL.
\end{abstract}


\section{Introduction}
\IEEEPARstart{T}{he} ability to learn from a limited number of examples and continuously acquire new knowledge without forgetting previously learned information is fundamental to intelligent systems. However, this capability remains a significant challenge for machine learning models, particularly in the context of Few-Shot Class-Incremental Learning (FSCIL). Traditional methods often suffer from catastrophic forgetting and over-fitting problems. Catastrophic forgetting means that the introduction of new classes leads to a severe degradation of performance on previously learned tasks. While over-fitting means that the model performs well on the training data but has poor generalization performance on new, unseen data, which arises because limited data may not fully represent the true distribution of the entire data generation process.

Despite of recent advances in deep learning have shown promise in addressing these challenges through various strategies, such as replay, meta-learning, and dynamic architectures, the lack of generalizable feature and the inability to better mimic human-like learning processes with limited data restrict further improvement in FSCIL performance.

Human-brain learning is capable of efficient adapting to novel concepts with very few examples. Biologists have tried to identify a number of underlying mechanisms that support continual learning with limited data. Some typical biological mechanisms include varying abstraction model for categorization \cite{opfer2005varying} and analogical learning mechanism \cite{gentner2017analogy}. The varying abstraction model represents a category through several clusters. Each cluster captures a distinct subset of features or characteristics associated with the category. This multi-cluster representation allows for greater flexibility and depth in understanding complex concepts. Analogical learning mechanism \cite{gentner2017analogy} forms target knowledge by analogizing the source knowledge to the target in some aspects. For instance, when learning to identify a zebra, we might draw comparisons with related concepts stored in our memory, such as horses, tigers, and pandas. Through this analogy, we note that a zebra shares certain features with each: it has a body structure reminiscent of a horse, exhibits the black-and-white coloration seen in pandas, and possesses stripes similar to those of a tiger. This analogical thinking helps us construct a new understanding of what defines a zebra.

Drawing inspiration from the brain's sophisticated mechanisms for categorization \cite{opfer2005varying} and analogical learning \cite{gentner2017analogy}, we propose Brain-inspired Analogical Mixture Prototypes (BAMP), a novel framework tailored for FSCIL scenarios. BAMP integrates mixed prototypical feature learning, statistical analogy, and soft voting to enable effective learning from small datasets while mitigating catastrophic forgetting and over-fitting. This method not only enhances the model's adaptability but also ensures robustness across diverse domains.

To evaluate the efficacy of BAMP, we conducted comprehensive experiments on six benchmark datasets: CIFAR100 \cite{krizhevsky2009learning}, CUB200 \cite{wah2011caltech}, EuroSAT \cite{helber2019eurosat}, FGVCAircraft \cite{maji2013fine}, Resisc-45 \cite{cheng2017remote} and StanfordCars \cite{krause20133d}. Our findings demonstrate that BAMP achieves state-of-the-art performance on both traditional big start FSCIL setting and challenging small start FSCIL setting. 

The main contributions of this paper are summarized as follows: 
\begin{enumerate}    
    \item A Brain-Inspired Framework for FSCIL: We propose Brain-inspired Analogical Mixture Prototypes (BAMP), a novel approach for Few-Shot Class-Incremental Learning (FSCIL) that integrates mixed prototypical feature learning, statistical analogy, and soft voting. This framework is inspired by the brain's mechanisms for categorization and analogical learning, and it effectively mitigates catastrophic forgetting and over-fitting while learning from limited data.
    \item Algorithmic Innovation: We develop a comprehensive algorithm for BAMP that includes model adaptation with mixed prototypical feature learning and incremental testing with statistical analogy and soft voting. This algorithm enables efficient learning from small datasets and ensures robust performance across diverse FSCIL scenarios.
    \item Superior Performance and Generalization: Through extensive experimentation and analysis on six benchmark datasets under both traditional big start and challenging small start FSCIL settings, BAMP demonstrates state-of-the-art performance by outperforming existing methods in terms of classification accuracy in the last session and average of the incremental accuracy.    
\end{enumerate}

The structure of this article is outlined as follows. In Section \ref{sec:relatedwork} we give a brief overview of the related work. In Section \ref{sec:problem_formulation}, we describe the problem formulation including definition of FSCIL and testing strategy. In Section \ref{sec:methods}, we present the details of our proposed BAMP. In Section \ref{sec:experiments} the datasets for training and evaluation, implementation details, experimental results, and ablation studies are presented. Finally, we summarize the conclusions in
Section \ref{sec:conclusions}.

\section{Related Work}
\label{sec:relatedwork}
Our approach involves FSCIL with pre-trained models and brain-inspired strategies. In this section, we describe the recent FSCIL methods with from-scratch models, FSCIL with pre-trained models, and brain-inspired FSCIL methods.
\subsection{FSCIL with From-Scratch Models}
FSCIL with from-scratch models usually apply small-scale models (such as ResNet18) as the backbone and train them from scratch. Existing methods can be categorized into data-driven, structure-based, and optimization-oriented approaches, with noted overlaps among these areas. For more comprehensive review, refer to the survey papers \cite{tian2024survey}, \cite{zhang2025few}. 

Data-driven approaches address FSCIL challenges from the data perspective. Relevant methodologies include data replay and pseudo-data construction. Data replay methods involve the storage \cite{kukleva2021generalized},\cite{zhu2022feature} or generation \cite{liu2022few},\cite{shankarampeta2021few} of limited amounts of old-class data. Pseudo-data construction  based methods generate synthetic classes and their corresponding samples to facilitate FSCIL models preparing for the real
incremental classes \cite{zhou2022forward},\cite{peng2022few}, or use base sessions to create pseudo-incremental sessions and meta-learning techniques to allow FSCIL models to understand how to handle incremental sessions \cite{zhang2021few},\cite{zhu2021self}.

Structure-based approaches leverage the architectural design or distinctive features of a model to tackle the challenges encountered in FSCIL. Related methods consist of dynamic structure methods and attention-based methods. The former seeks to accomplish FSCIL by adaptively modifying the model's architecture or the relationships among prototypes \cite{tao2020few},\cite{yang2022dynamic}. While the latter modifies the feature attention allocation by integrating attention modules into the model architecture \cite{zhao2023few},\cite{zhou2022few}, thus allows the model to focus on information relevant to the current task, improving its performance and generalization ability.

Optimization-oriented approaches address the challenges in FSCIL by tackling the complexities of optimization problems. Key strategies include representation learning, meta-learning, and knowledge distillation (KD). Representation learning focuses on deriving meaningful features from a restricted flow of samples to construct a "representation" of the data \cite{mazumder2021few},\cite{hersche2022constrained}. Meta-learning utilizes insights gathered from numerous learning episodes, covering a distribution of related tasks, to improve future learning efficiency \cite{zheng2021few},\cite{zhou2022few}. KD focuses on transferring knowledge between sessions \cite{dong2021few},\cite{cheraghian2021semantic}. 

Different from methods with from-scratch models, our approach is based on pre-trained models and fine-tune them with  brain-inspired technique to address the challenges of FSCIL.
\subsection{FSCIL with Pre-Trained Models}
Pre-trained models, such as Vision Transformers (ViT), have shown significant potential in FSCIL. These models are pre-trained on large datasets and fine-tuned for specific tasks. Recent studies have demonstrated that leveraging pre-trained models can significantly improve performance in FSCIL scenarios. 

FeCAM \cite{goswami2024fecam} utilizes frozen pre-trained model as feature extractor to generate new class prototypes, and classify the features based on the anisotropic Mahalanobis distance to the prototypes. C-FeCAM \cite{goswami2024calibrating} complements FeCAM \cite{goswami2024fecam} by utilizing calibrated prototypes and covariances to improve classification performance. RanPAC \cite{mcdonnell2024ranpac} utilizes random projections combined with non-linear activations to transform pre-trained model's features into an extremely high-dimensional space, thereby enhancing their linear separability for improved classification in this expanded space. C-RanPAC \cite{goswami2024calibrating} is the calibrated version of RanPAC, with a calibration process similar to that of C-FeCAM \cite{goswami2024calibrating}. Qiu et al. \cite{qiu2023semantic} have introduced a semantic-visual guided transformer (SV-T) aimed at boosting the feature extraction capabilities of pre-trained backbones for incremental class learning. Meanwhile, Park et al \cite{park2024pre} propose PriViLege, which leverages the Vision Transformer (ViT) architecture, to adeptly address the issues of catastrophic forgetting and overfitting in large-scale models by utilizing pretrained knowledge tuning (PKT). These methods are all based on pre-trained models and fine-tune them with specific techniques to address the challenges of FSCIL. 

Our method is also based on pre-trained ViT models, however, we focus on the fine-tune strategy and the prototype calibration technique inspired by the brain's mechanisms. 
\subsection{Brain-Inspired FSCIL}
Currently, there is scant research focusing on brain-inspired FSCIL. Wang et al. \cite{wang2021few} introduce a brain-inspired two-step consolidation approach to FSCIL that prevents forgetting while enhancing generalization without overfitting. Zhao et al. \cite{zhao2021mgsvf} propose a strategy called SvF, grounded in the fast and slow learning mechanism, designed to preserve existing knowledge and swiftly acquire new information. Ran et al. \cite{ran2024brain} present a brain-inspired prompt tuning method, i.e., fast- and slow-update prompt tuning FSCIL (FSPT-FSCIL), for transferring foundation models to the FSCIL task. 

Different from these methods, our approach draws inspiration from the brain's mechanisms for categorization and analogical learning, fine-tunes pre-trained Vision Transformer (ViT) with mixed prototypical feature learning in base session, and analogically calibrate statistics of prototypes at incremental testing stage.
\section{Problem formulation}
\label{sec:problem_formulation}
We first introduce the FSCIL learning paradigm and then present the testing strategy.
\subsection{The Problem Definition}
In the FSCIL (Few-Shot Class-Incremental Learning) problem, we have a sequence of labeled training sets \(\{\mathcal{D}^0, \mathcal{D}^1, \mathcal{D}^2, \ldots, \mathcal{D}^{n-1}\}\), test sets \(\{\mathcal{T}^0, \mathcal{T}^1, \mathcal{T}^2, \ldots, \mathcal{T}^{n-1}\}\), and class label sets \(\{\mathcal{C}^0, \mathcal{C}^1, \mathcal{C}^2, \ldots, \mathcal{C}^{n-1}\}\). Here, $n$ is the total number of sessions or tasks. Each training set \(\mathcal{D}^t\) consists of labeled examples \(\{(\mathbf{x}_i^t, y_i^t)\}_{i=1}^{|\mathcal{D}^t|}\), where \(y_i^t\) belongs to \(\mathcal{C}^t\). The class label sets are mutually exclusive, meaning \(\mathcal{C}^i \cap \mathcal{C}^j = \emptyset\) for all \(i \neq j\). The training set \(\mathcal{D}^0\) of the first session contains a sufficient number of examples for each class in \(\mathcal{C}^0\). For subsequent sessions \(t > 0\), \(\mathcal{D}^t\) is an \(N\)-way \(K\)-shot training set, which includes \(N\) classes with \(K\) examples per class from \(\mathcal{C}^t\). At the end of each session \(t\), the model is evaluated using a combined test set \(\mathcal{T}_{0 \sim t} = \mathcal{T}^0 \cup \cdots \cup \mathcal{T}^t\).

\subsection{Testing Strategy} 
Follow commonly setting in community, we test BAMP in the context of both traditional big start setting and the challenging small start setting on six FSCIL benchmark datasets. Training of both settings uses all available samples for each class in the many-shot base task, and $5$ samples only from each class in the few-shot tasks, i.e., $5$-shot setting. In the big start setting, $50\%$ of the classes are used in the first task, and the other $50\%$ of the classes are split equally in the incremental few-shot tasks. For the small start setting, the dataset is equally split in all tasks. Performance on each dataset is measured in terms of classification accuracy for all seen classes after learning in the last session $A_{last}$, and the average of the incremental accuracy from all sessions $A_{inc}$. Overall performance is measured in terms of average of $A_{last}$ and $A_{inc}$ over 6 datasets, i.e., $mA_{last}$ and $mA_{inc}$. Details of datasets and settings are described in Section \ref{sec:methods}.

\section{Methods}
\label{sec:methods}
\subsection{The framework of Brain-inspired Analogical Mixture Prototypes (BAMP)}
\label{subsec:framework}
\begin{figure*}[h]  
\centering
\setlength{\abovecaptionskip}{-2.5cm}
\includegraphics[width=\textwidth]{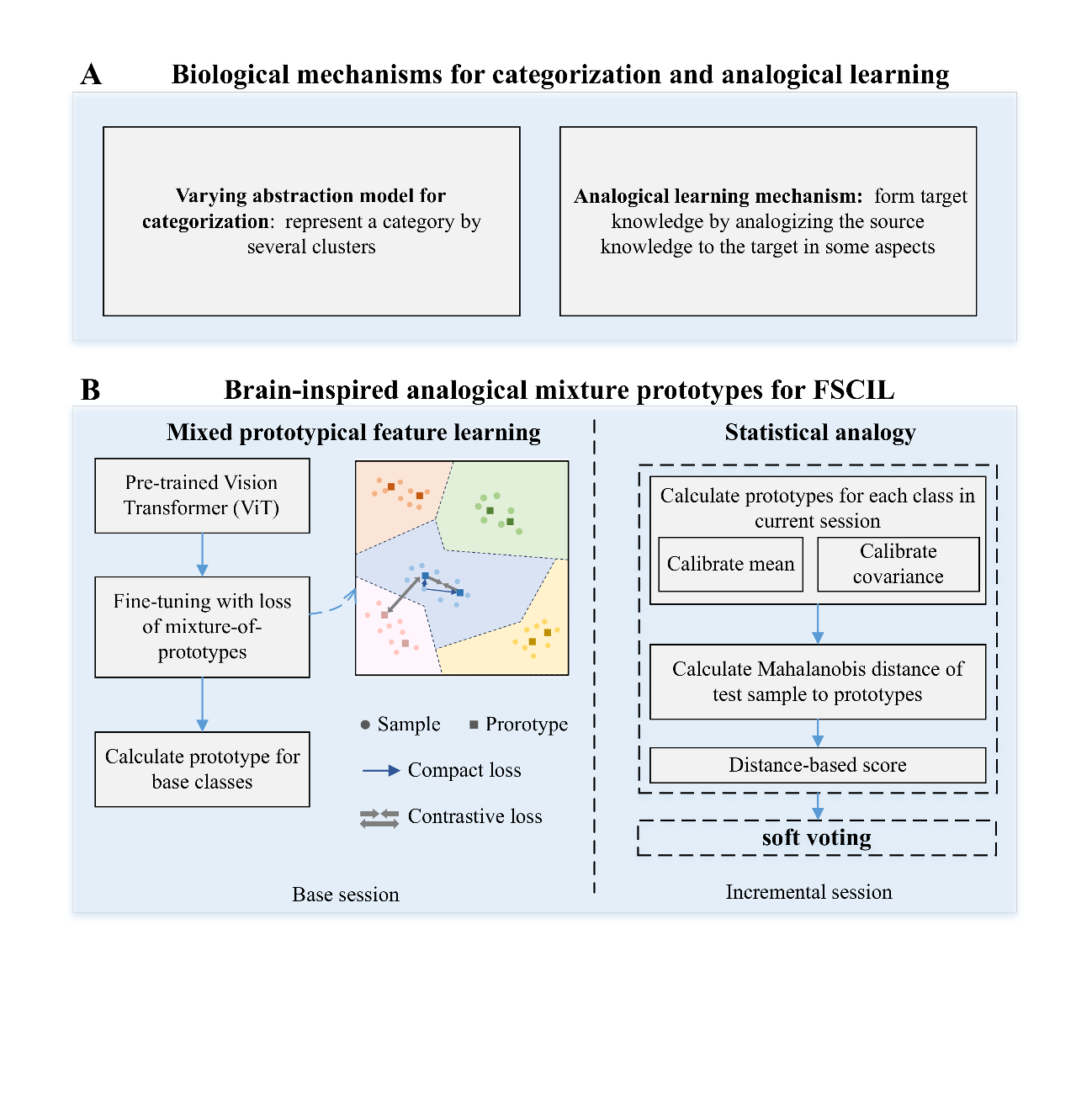}
\caption{Schematic of the brain-inspired analogical mixture prototypes. \textbf{A} The biological mechanisms of categorization and analogical learning. \textbf{B} The brain-inspired analogical mixture prototypes for FSCIL.}
\label{fig:framework}
\end{figure*}

The framework of brain-inspired analogical mixture prototypes is shown in Figure \ref{fig:framework}. There are three components in the framework, i.e., mixed prototypical feature learning, statistical analogy, and soft voting. We will describe the role of each component briefly below. 
\begin{itemize}
    \item Mixed Prototypical Feature Learning. It is utilized to learning representative feature in the base session. Firstly, started from a pre-trained Vision Transformer (ViT), the mixed prototypical feature learning represents each class with a mixture of prototypes to capture variability within each class. Secondly, it fine-tunes the ViT model with compact loss and contrastive loss, helping to encourage the sample embeddings to be compact around their associated prototypes, and also encourage intra-class compactness and inter-class discrimination at the prototype level. Thirdly, it computes a representative feature vector (prototype) for each base class, which is used for classification, ensuring that the model has a clear and compact representation of each class.     
    \item Statistical Analogy: This component is used to form concept of incremental sessions from limited samples by analogy. It firstly computes the prototypes for new classes in the current session, ensuring that the model can represent new classes effectively by capturing their characteristic features. And then calibrates the mean and covariance matrix of prototypes for new classes according to their similarity to base classes. Finally, it computes classification scores using Mahalanobis distance, which helps in distinguishing between classes more effectively. This statistical analogy ensures that the model can leverage the knowledge from previously learned classes to better understand and classify new classes.    
    \item Soft Voting: This component combines the merits of statistical analogy with an off-the-shelf FSCIL method, enhancing the robustness and accuracy of the model. Soft voting allows the model to make more informed decisions by integrating multiple sources of information, thereby improving its overall performance.
\end{itemize}

\subsection{BAMP Algorithm}
\label{subsec:methods_BAMP_algorithm}
BAMP algorithm consists two stages: (a) model adaptation with mixed prototypical feature learning and (b) incremental testing by statistical analogy and soft voting. We provide the pseudo code in Algorithm \ref{algo:BAMP}. The more information about stages of the Brain-inspired Analogical Mixture Prototypes algorithm (BAMP) are described below.

\begin{algorithm}
\caption{Brain-inspired analogical mixture prototypes (BAMP) for FSCIL}\label{algo:BAMP}
\begin{algorithmic}[1]
\Require Incremental training sets: $\{\mathcal{D}^0, \mathcal{D}^1, \cdots, \mathcal{D}^{n-1}\}$; Incremental test sets: $\{\mathcal{T}^0, \mathcal{T}^1, \mathcal{T}^2, \ldots, \mathcal{T}^{n-1}\}$; Pre-trained Model: $f(\mathbf{x})$
\Ensure Updated model $f^*(\mathbf{x})$; Classification results \(\{R^{0}, R^{1}, \ldots, R^{n-1}\}\)
\State $f^*(\mathbf{x})\leftarrow$ Adapt the model to $\mathcal{D}^0$ via Eq. \ref{eq:modelAapt} and Eq. \ref{equ:l_bamp} \Comment{Model adaptation}
\State Freeze the embedding functions $\phi^*(\cdot)$
\State Extract prototypes for base classes via Eq. \ref{eq:proto_base}
\State Init the combined test set: $\mathcal{T} = \emptyset$
\For{$t = 1, 2, \cdots, n-1$} \Comment{Incremental testing}
    \State Get the incremental training set $\mathcal{D}^t$
    \State Get the incremental test set $\mathcal{T}^t$
    \State Update the combined test set, $\mathcal{T} = \mathcal{T} \cup \mathcal{T}^t$
    \State Extract the prototypes of training set $\mathcal{D}^t$ for current session via Eq. \ref{eq:proto_crt_session}
    \State Perform statistical analogy for test samples in $\mathcal{T}$ via Eq. \ref{eq:sim}-\ref{eq:minMaxNorm}
    \State Perform soft voting for test samples in $\mathcal{T}$ via Eq. \ref{eq:softVoting}       
    \State Get the predictions of the combined test set $\mathcal{T}$  via Eq. \ref{eq:finalClassfier}
    \State $R^{t}\leftarrow$ Calculate the classification accuracy of combined test set $\mathbf{T}$ 
\EndFor
\State \Return the updated model $f^*(\mathbf{x})$ and classification results \(\{R^{0}, R^{1}, \ldots, R^{n-1}\}\)
\end{algorithmic}
\end{algorithm}

\subsubsection{Model Adaptation}
 Starting from a pre-trained ViT \cite{dosovitskiy2020image}, we adapt the model in the base session with mixed prototypical feature learning. For the network architecture, we follow \cite{ijcv_2024_Revisiting} to integrate an adapter into the original ViT's MLP structure. The integrated adapter is a bottleneck module comprising a down-projection \( W_{\text{down}} \in \mathbb{R}^{d \times r} \) for reducing the feature dimension, a non-linear activation function, and an up-projection \( W_{\text{up}} \in \mathbb{R}^{r \times d} \) to restore the original dimension. $\theta=\theta_{W_{\text{down}}} \cup \theta_{W_{\text{up}}}$ is parameter set of the adapter. For clarity, we decompose the deep model into two parts: \( f(\mathbf{x}) = W^\top \phi(\mathbf{x}) \), where \( \phi(\cdot) : \mathbb{R}^D \to \mathbb{R}^d \) is the embedding function and \( W \in \mathbb{R}^{d \times |Y_k|} \) serves as the linear classification head. The model adaptation process on the base session can be formulated as: 
\begin{equation}
f^*(\mathbf{x}) = \mathcal{F}(f(\mathbf{x}), \mathcal{D}^0, \Theta).
\label{eq:modelAapt}
\end{equation}
This equation indicates that the adapted model \( f^*(\mathbf{x}) \) is a function of the original model \( f(\mathbf{x}) \), the first incremental dataset \( \mathcal{D}^0 \), and the parameter set \( \Theta \). With pre-trained weights of ViT frozen, the adaptation process optimizes the adapter and classification head, i.e., $\Theta = \theta_{W_{\text{down}}} \cup \theta_{W_{\text{up}}} \cup \theta_W$. The corresponding adapted embedding function is denoted by $\phi^*(\cdot,\theta)$. 
 
 For the adaptation, we differs from \cite{ijcv_2024_Revisiting} in class representation and the loss function. We represent each class using multiple prototypes, each modeled with a mixture of von Mises-Fisher (vMF) distributions, motivated by \cite{lu2024learning_iclr}, which utilizes Mixture of prototypes for Out-of-distribution (OOD) detection. Additionally, our loss function comprises three items: cross-entropy for the linear classification head, compact loss, and contrastive loss for the mixture-of-prototypes representation. While \cite{ijcv_2024_Revisiting} represents a class by a prototype and use the cross entropy loss only. The process for assignment, pruning and updating of prototypes is same as \cite{lu2024learning_iclr}. 

 \textbf{Embedding space modeling}. Similar to \cite{lu2024learning_iclr}, we formulate the embedding space using a hyperspherical model. Projected embeddings \(\mathbf{z}\) that lie on the unit sphere (\(\|\mathbf{z}\|^2 = 1\)) can be naturally modeled using the von Mises-Fisher (vMF) distribution. The entire embedding space is modeled as a mixture of vMF distributions \cite{mardia2000directional,wang2020understanding}, where each distribution is defined by a mean \(\mathbf{p}_k\) and a concentration parameter \(\kappa\):
 \begin{equation}
 p_d(\mathbf{z}; \mathbf{p}_k, \kappa) = Z_d(\kappa) \exp \left( \kappa \langle \mathbf{p}_k, \mathbf{z} \rangle \right),  
\label{eq:vMF}
\end{equation}
where \(\mathbf{p}_k \in \mathbb{R}^d\) is the \(k\)-th prototype with unit norm, \(\kappa \geq 0\) represents the tightness around the mean, and \(Z_d(\kappa)\) is the normalization factor,  $\langle \cdot, \cdot \rangle$ is inner product of two vectors. 

 \textbf{Class representation}. We use the embedding function \( \phi(\cdot) \)  to encode and normalize the input \( \mathbf{x} \) to lower-dimensional embedding \( \mathbf{z} \), i.e. \( \mathbf{z} = \phi(\mathbf{x}) \) , and model each class using a mixture of vMF distributions with multiple prototypes. For each class \( c \), we define \( K \) prototypes \( \mathbf{P}^c = \{\mathbf{p}_k^c\}_{k=1}^K \), corresponding to a mixture of vMF distributions.  Each sample \( \mathbf{z}_i \) is assigned to these prototypes via assignment weights \( \mathbf{w}_i^c \in \mathbb{R}^K \). The probability density for a sample embedding \( \mathbf{z}_i \) in class \( c \) is defined as:
\begin{equation}
p(\mathbf{z}_i; \mathbf{w}_i^c, \mathbf{P}^c, \kappa) = \sum_{k=1}^K w_{i,k}^c p_d(\mathbf{z}_i; \mathbf{p}_k^c, \kappa)
\label{eq:mixturemodel}
\end{equation}

where \( w_{i,k}^c \) denotes the \( k \)-th element of \( \mathbf{w}_i^c \). The same number of prototypes, \( K \), are defined for each class. The assignment weights are determined based on the adjacency relationship between the samples and the estimated prototypes. Given the probability model in Eq. \ref{eq:mixturemodel}, prototype-based classifier assigns an embedding \( \mathbf{z}_i \) to a class \( c \) with the following normalized probability:

{
\small
\begin{equation}
p(y_i = c | \mathbf{z}_i; \{\mathbf{w}_i^j, \mathbf{P}^j, \kappa\}_{j=1}^C) = \frac{\sum_{k=1}^K w_{i,k}^c \exp(
\langle \mathbf{p}_k^c, \mathbf{z}_i \rangle / \tau)}{\sum_{j=1}^C \sum_{k'=1}^K w_{i,k'}^j \exp( \langle \mathbf{p}_{k'}^j, \mathbf{z}_i \rangle   / \tau)},
\label{eq:prob}
\end{equation}
}
where \( \tau = 1/\kappa \) is analogous to the temperature parameter in the compact loss.

 \textbf{Training objectives}.
The overall training objective of BAMP is defined as:
\begin{equation} 
\mathcal{L}_{\text{BAMP}} =\mathcal{L}_{\text{CE}} + \alpha \mathcal{L}_{\text{com}} + \lambda \mathcal{L}_{\text{proto-contra}},
\label{equ:l_bamp}
\end{equation}
where \( \mathcal{L}_{\text{CE}} \) is the cross entropy loss, \( \mathcal{L}_{\text{com}} \) represents the compact loss and    \( \mathcal{L}_{\text{proto-contra}} \) indicates the contrastive loss, $\alpha > 0$ and $\lambda > 0$ are weights to control the balance among these three loss functions.

We use cross-entropy loss to evaluate the performance of the linear classification head, which outputs a probability value between 0 and 1. The loss increases as the predicted probability deviates from the actual label. The \( \mathcal{L}_{\text{CE}} \) is defined as:

\begin{equation} 
\mathcal{L}_{\text{CE}} = -\frac{1}{N} \sum_{i=1}^{N} \sum_{c=1}^{C} y_i^c \log(\hat{y}_i^c),
\label{equ:ce_loss}
\end{equation}
where \(y_i^c\) is the binary indicator (0 or 1) of whether class label \(c\) is the correct classification for the current training example $\mathbf{x}_i$, \(\hat{y}_i^c\) is the predicted probability that the current training example $\mathbf{x}_i$ belongs to class \(c\), \(N\) is the total number of training examples. 

In term of compact loss, to promotes compactness between samples and prototypes, we perform maximum likelihood estimation (MLE) on the training data, according to Eq. \ref{eq:prob} by solving the problem:

\begin{equation}
\max_{\theta} \prod_{i=1}^N p(y_i = c | \mathbf{z}_i, \{\mathbf{w}_i^c, \{\mathbf{p}_k^c, \kappa\}_{k=1}^K\}_{c=1}^C),
\label{equ:problem2solve}
\end{equation}

where \(\mathbf{z}_i\) is the hyperspherical embedding, \(\theta\) is parameter set of the adapter. By taking the negative log-likelihood, the optimization problem can be rewritten as the compact loss \( \mathcal{L}_{\text{com}} \):

\begin{equation} 
 \mathcal{L}_{\text{com}} =  -\frac{1}{N} \sum_{i=1}^N \log \frac{\sum_{k=1}^K w_{i,k}^{y_i} \exp ( \langle \mathbf{p}_k^{y_i}, \mathbf{z}_i \rangle / \tau)}{\sum_{c=1}^C \sum_{k'=1}^K w_{i,k'}^c \exp (\langle\mathbf{p}_{k'}^c, \mathbf{z}_i\rangle / \tau)},
\label{eq:mle}
\end{equation}

where \(y_i\) represents the class index of sample embedding \(\mathbf{z}_i\), and \(\tau\) is the temperature parameter. This compact loss for MLE encourages samples to be close to their respective prototypes.

The prototype contrastive loss relies on the class information of the prototypes as an implicit supervision signal. The loss encourages intra-class compactness and inter-class discrimination at the prototype level, thus the embedding space is further refined. The contrastive loss is defined as:

\vspace{15pt}
\begin{strip}
\begin{equation}
\mathcal{L}_{\text{proto-contra}} = -\frac{1}{CK} \sum_{c=1}^C \sum_{k=1}^K \log \frac{\sum_{k'=1}^K \mathbbm{1}(k' \neq k) \exp (\langle \mathbf{p}_k^c, \mathbf{p}_{k'}^c\rangle / \tau)}{\sum_{c'=1}^C \sum_{k''=1}^K \mathbbm{1}(k'' \neq k, c' \neq c) \exp (\langle \mathbf{p}_k^c, \mathbf{p}_{k''}^{c'}\rangle
/ \tau)},
\label{eq:protoContra}
\end{equation}
\end{strip}
where \(\mathbbm{1}(\cdot)\) is an indicator function to avoid contrasting the same prototype. The prototypes are updated using the exponential moving average (EMA) technique. Thus, the loss in Eq. \ref{eq:protoContra} primarily regularizes the sample embeddings through the connection between samples and prototypes.

\subsubsection{Incremental Testing}
When the model adaptation stage is done, we extract prototypes for base classes with the embedding function \( \phi^*(\cdot) \) of the trained model immediately as:
 \begin{equation}
     \mathbf{p}^c = \frac{1}{M} \sum_{i=1}^{|\mathcal{D}^0|} \mathbb{I}(y_i = c) \phi^*(\mathbf{x}_i),
 \label{eq:proto_base}
 \end{equation} 
 where \( M = \sum_{i=1}^{|\mathcal{D}^0|} \mathbb{I}(y_i = c) \), and \( \mathbb{I}(\cdot) \) is the indicator function. The averaged embedding captures the most common pattern of the corresponding class \(c\). In each incremental session, we conduct incremental testing by statistical analogy and soft voting. Below we give the detailed breakdown of incremental testing. 

 \textbf{Statistical analogy} forms class concept of incremental sessions from limited samples by analogy.  We provide details of statistical analogy as follows. It firstly calculates the prototypes for each new class in the current session as:
  \begin{equation}
     \mathbf{P}^c = \{\mathbf{p}_k^c\}_{k=0}^n,
 \label{eq:proto_crt_session}
 \end{equation}  
 where \( n \) is the number of samples for class \(c\) and usually a small integer, for example \( 5 \) corresponds to 5-shot setting, \( \mathbf{p}_0^c \) is the mean feature vector of n-shot samples for class \(c\), and the rest prototypes are the feature embeddings of the n-shot samples $\{\mathbf{x}_k^c \}_{k=1}^n$, i.e. \(\{\mathbf{p}_k^c\}_{k=1}^n = \{\mathbf{z}_k^c|\mathbf{z}_k^c=\phi^*(\mathbf{x}_k^c) \}_{k=1}^n\). 
 
 And then, we calibrate the mean and covariance of prototypes for new classes according to their similarity to base classes. The similarity between a base class prototype \(\mathbf{p}^b\) and a prototype \(\mathbf{p}_k^c\) of new class \(c\) can be used to compute weights for averaging old class statistics with new ones. We calculate the cosine similarity \(S_{b,ck}\) between \(\mathbf{p}^b\) and \(\mathbf{p}_k^c\) as follows:

\begin{equation}
S_{b,ck} = \frac{ \langle \mathbf{p}^b, \mathbf{p}_k^c\rangle}{\|\mathbf{p}^b\| \cdot \|\mathbf{p}_k^c\|} \cdot \tau,
\label{eq:sim}
\end{equation}

where \(\tau\) controls the sharpness of the weight distribution. Following \cite{wang2024few}, we set \(\tau = 16\) in our experiments.

The weight \(w_{b,ck}\) of the new class prototype \(\mathbf{p}_k^c\) corresponding to a base class prototype \(\mathbf{p}^b\) is obtained by performing softmax over all base class prototypes:

\begin{equation}
w_{b,ck} = \frac{e^{S_{b,ck}}}{\sum_{i=1}^{B} e^{S_{i,ck}}},
\label{eq:weight}
\end{equation}

such that \(\sum_{b=1}^{B} w_{b,ck} = 1\) for a prototype \(\mathbf{p}_k^c\) of new class \(c\), where \(B\) is the number of base classes.
 
Mean calibration. Similar to \cite{wang2024few}, the biased prototype means of new classes \(\mathbf{p}_k^c\) can be calibrated using the following formula:
\begin{equation}
\hat{\mathbf{p}}_k^c = \beta \mathbf{p}_k^c + (1 - \beta) \sum_{b=1}^{B} w_{b,ck} \mathbf{p}^b,
\label{eq:meanCalib}
\end{equation}
here, \(\beta\) controls the degree of calibration.
 
Covariance calibration. Similar to \cite{goswami2024calibrating}, we use the softmaxed similarity weights from Eq. \ref{eq:weight} to calibrate the new class covariances \(\Sigma_k^c\) by averaging with the base class covariances \(\Sigma_b\):
\begin{equation}
     \hat{\Sigma}_k^c = \eta \left( \Sigma_k^c + \sum_{b=1}^{B} w_{b,ck} \Sigma_b \right),
 \label{eq:calibrated_cov}
 \end{equation}  
here, \(\eta\) controls the scaling of the covariance matrix.
 
 Finally, we compute statistical analogy-based classification score using Mahalanobis distance. The details are as follows. To obtain an invertible full-rank covariance matrix, we perform covariance shrinkage \cite{goswami2024fecam}. By performing correlation normalization \cite{goswami2024fecam}, the correlation matrix is calculated from the shrunk covariance matrix. We compute the Mahalanobis distance between the prototypes and the test features using the correlation matrix of each prototype as follows:
 {
 \small
\begin{equation}
\mathcal{D}_M(\phi^*(x), \hat{\mathbf{p}}_k^c) = (\phi^*(x) - \hat{\mathbf{p}}_k^c)^T \left( N(\hat{\Sigma}_k^c + \gamma I) \right)^{-1} (\phi^*(x) - \hat{\mathbf{p}}_k^c).
\label{eq:mahaDist}
\end{equation}
}

Here, \(\hat{\mathbf{p}}_k^c\) denotes the calibrated prototypes from Eq. \ref{eq:meanCalib}, \(\hat{\Sigma}_k^c\) represents the calibrated covariances obtained from Eq. \ref{eq:calibrated_cov}, \(\phi^*(x)\) is the feature vector extracted from the test samples, and \(N\) indicates the correlation normalization. For base classes, we set \(\hat{\Sigma}_k^c = \Sigma_k^c\). The statistical analogy-based classification score of test sample \(x\) with respect to class \(c\) is defined as:
\begin{equation}
sc_{sa}(\phi^*(x), c)= \max_{1 \leq k \leq K} \exp(- \mathcal{D}_M(\phi^*(x), \hat{\mathbf{p}}_k^c) ).
\label{eq:scoreOfStat}
\end{equation}
The score vector of test sample \(x\) with respect to all seen classes is formulated as:
 \begin{equation}
 \mathbf{s}_{sa}(\phi(x)) = N_{\text{minmax}}((sc_{sa}(\phi(x), c))_{c=1}^C),
 \label{eq:scorevecOfStat}
 \end{equation}  
 where \(K\) is the number of prototypes for class \(c\), \(C\) is the total number of seen classes. $N_{\text{minmax}}$ is the Min-Max Normalization technique that scales the range of $\mathbf{s}_{sa}$ to [0, 1], ensuring that the statistical analogy-based classification scores can be compared with those from other methods. The Min-Max Normalization is formulated as:
  \begin{equation}
 N_{\text{minmax}}(\mathbf{x}) = \frac{\mathbf{x} - \min(\mathbf{x}) }{\max(\mathbf{x}) - \min(\mathbf{x})}
 \label{eq:minMaxNorm}
 \end{equation}  
where $\min(\mathbf{x})$ and $\max(\mathbf{x})$ are the minimum and maximum elements of the vector $\mathbf{x}$, respectively.

 \textbf{Soft voting} integrates the strengths of both statistical analogy and an off-the-shelf FSCIL method to further improve overall performance. Specifically, soft voting aggregates the probability estimated from statistical analogy and an off-the-shelf FSCIL classifier:

\begin{equation}
 \mathbf{s} = \mathbf{s}_{sa}(\phi(x)) + \mathbf{s}_{ots}(\phi(x)),
 \label{eq:softVoting}
 \end{equation}
 where $\mathbf{s}_{sa}(\phi(x))$ and $\mathbf{s}_{ots}(\phi(x))$ are the probabilities from the statistical analogy and the off-the-shelf FSCIL classifier, respectively. And the class with the highest average probability is selected as the final prediction:
\begin{equation}
 \hat{y}=  \arg\max_j \mathbf{s}_j,
 \label{eq:finalClassfier}
 \end{equation}

\section{Experimental results}
\label{sec:experiments}
\subsection{Experimental Setup}
\subsubsection{Datasets and Task Protocols}
\label{subsec:methods_datasets}
The experiments are conducted on six publicly available datasets, including  CIFAR100 \cite{krizhevsky2009learning}, CUB200 \cite{wah2011caltech}, EuroSAT \cite{helber2019eurosat}, FGVCAircraft \cite{maji2013fine}, Resisc-45 \cite{cheng2017remote} and StanfordCars \cite{krause20133d}. The CIFAR-100 dataset \cite{krizhevsky2009learning} comprises 50,000 images designated for training and 10,000 images set aside for testing, with these images distributed across 100 distinct categories. The CUB-200 dataset \cite{wah2011caltech} features 5,994 images intended for training purposes and 5,794 images for testing, spanning 200 different bird species. The EuroSAT Dataset \cite{helber2019eurosat} includes 27,000 labeled RGB image patches divided among 10 classes that represent different types of land use and land cover. FGVC-Aircraft \cite{maji2013fine} includes a total of 10,200 images that cover 102 types of aircraft variants. For our purposes, we have selected 100 classes at random from this dataset for use in our analysis. Resisc-45 \cite{cheng2017remote} contains 31,500 RGB images of size 256×256 divided into 45 scene classes, each class containing 700 images. StanfordCars \cite{krause20133d} encompasses a collection of 8,144 images that are utilized for both training and evaluation, representing 196 classes of car models.

We follow both the commonly used setting of having a big first task(50\% of the classes in the first task) as well as the challenging small start setting (equally split the dataset in all tasks). For training in the many-shot base task, we use all available samples for each class, while 5 samples only of each class in the incremental few-shot tasks. In Tab. \ref{tab:protocol}, we summarize the details of the total number of classes, number of sessions, number of classes in the base session (Base-m), and the number of classes in each incremental task (Inc-n) for all datasets.

\begin{table*}[htbp]
\centering
\caption{Details on class splits over the continual sessions for different datasets}\label{tab:protocol}%
\begin{tabular}{@{}l *{7}{c} @{}}
\toprule
\multirow{2}{*}{ Dataset}&\multirow{2}{*}{Classes}&\multicolumn{3}{@{}c@{}}{Big start setting} &\multicolumn{3}{@{}c@{}}{Small start setting} \\
\cmidrule{3-5}\cmidrule{6-8}%
 &   & Sessions & Base-m\footnotemark[1] & Inc-n\footnotemark[2]& Sessions & Base-m & Inc-n\\
\midrule
CIFAR100 \cite{krizhevsky2009learning}    & 100   & 6  & 50 & 10  & 10  & 10 & 10 \\
CUB200 \cite{wah2011caltech}    & 200   & 11  & 100 & 10 & 10  & 20 & 20 \\
EuroSAT \cite{helber2019eurosat}   & 10   & 6  & 5 & 1 & 5  & 2 & 2 \\
FGVCAircraft \cite{maji2013fine}    & 100   & 11  & 50 & 5 & 10 & 10 & 10 \\
Resisc-45 \cite{cheng2017remote}   & 45   & 6  & 20 & 5 & 9  & 5 & 5 \\
StanfordCars \cite{krause20133d}   & 196   & 8  & 98 & 14 & 7  & 28 & 28 \\
\bottomrule
\end{tabular}
\footnotetext[1]{Base-m stands for the number of classes in the base session.}
\footnotetext[2]{Inc-n stands for the number of classes in each incremental task.}
\end{table*}

\subsubsection{Implementation Details}
We use the ViT-B/16 model from the timm library \cite{timm_library}, pre-trained on ImageNet-21K and fine-tuned on ImageNet-1k, similar to \cite{ijcv_2024_Revisiting}. Following \cite{chen2022adaptformer,ijcv_2024_Revisiting}, we employ a ViT adaptor to adapt the pre-trained ViT to the dataset for the first session over 100 epochs. Intuitively, an increased number of classes facilitates the training of features that exhibit stronger contrast. So we set $\alpha$ and $\lambda$ of the training objective (Eq. \ref{equ:l_bamp}) to $\min(1.0, \exp(-(20/nbase-1)))$ according the total number of class of the base session $nbase$. 

For mean calibration (Eq. \ref{eq:meanCalib}) in C-FeCAM \cite{goswami2024calibrating}, C-RanPAC \cite{goswami2024calibrating} and BAMP, we set \(\beta = 0.75\) for CIFAR100 \cite{krizhevsky2009learning} and \(\beta = 0.9\) for other datasets including CUB200 \cite{wah2011caltech}, EuroSAT \cite{helber2019eurosat}, FGVCAircraft \cite{maji2013fine}, Resisc-45 \cite{cheng2017remote} and StanfordCars \cite{krause20133d}. For covariance calibration (Eq. \ref{eq:calibrated_cov}), we use \(\eta = 1\) for C-FeCAM \cite{goswami2024calibrating} and the proposed BAMP, and \(\eta = 0.5\) for C-RanPAC \cite{goswami2024calibrating} across all datasets. To compute the Mahalanobis distance, we use $\gamma=500$ in Eq. \ref{eq:mahaDist}. In soft voting stage, we use C-RanPAC \cite{goswami2024calibrating} as the off-the-shelf FSCIL classifier.

\subsubsection{Description of Compared Methods}
\label{subsec:methods_desc_compared_methods}
We compare our proposed approach with four typical FSCIL methods: FeCAM \cite{goswami2024fecam}, C-FeCAM \cite{goswami2024calibrating}, RanPAC \cite{mcdonnell2024ranpac} and C-RanPAC \cite{goswami2024calibrating}. FeCAM \cite{goswami2024fecam} employs prototypical networks and the anisotropic Mahalanobis distance to enhance classification performance. C-FeCAM \cite{goswami2024calibrating} complements FeCAM \cite{goswami2024fecam} by utilizing calibrated prototypes and covariances for each new class. RanPAC \cite{mcdonnell2024ranpac} utilizes random projections combined with non-linear activations to transform features into an extremely high-dimensional space, thereby enhancing their linear separability for improved classification in this expanded space. C-RanPAC \cite{goswami2024calibrating} is the calibrated version of RanPAC, with a calibration process similar to that of C-FeCAM \cite{goswami2024calibrating}.

\subsection{BAMP Outperforms State-of-the-art on Traditional Big Start FSCIL Setting}
\label{subsec:Results_big_start_setting}
\begin{figure*}[h]  
\centering
\includegraphics[width=\textwidth]{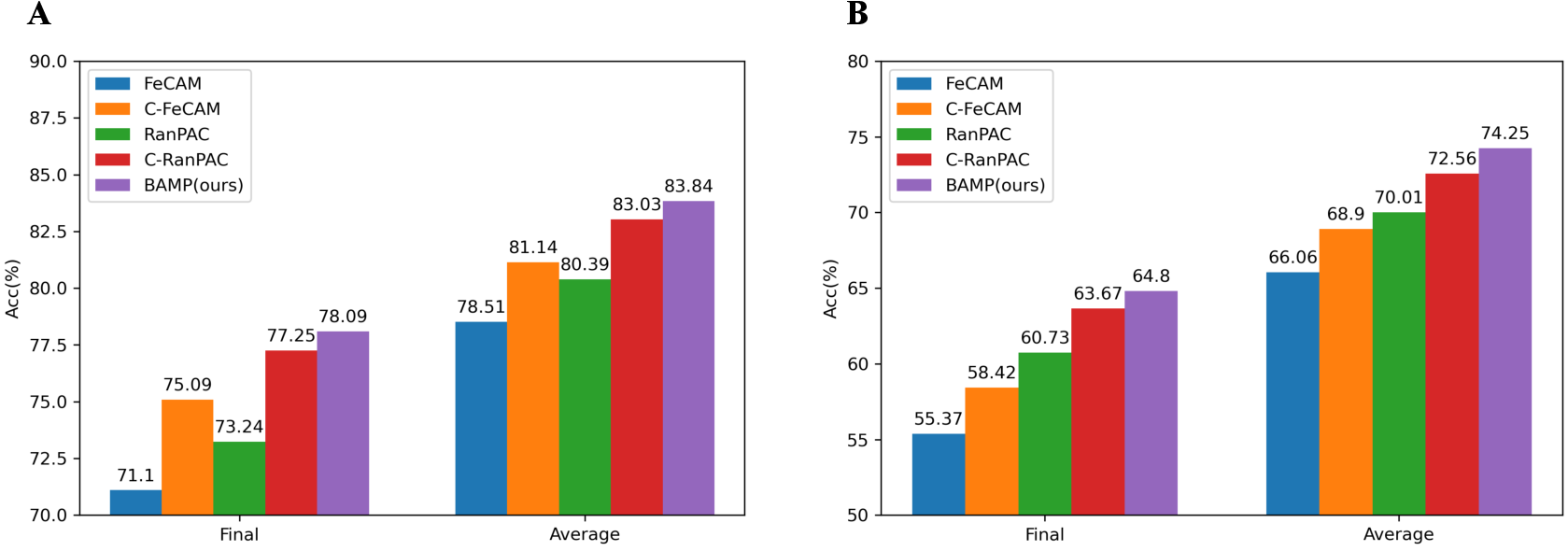}
\caption{Average of $A_{last}$ and $A_{inc}$ over 6 datasets. ``Final'' refers to the classification accuracy (ACC) for all seen classes in the last session. ``Average'' refers to the mean ACC across all sessions. \textbf{A} is for the big start setting, while \textbf{B} is for the small start setting.}
\label{fig:avg6ds}
\end{figure*}

We evaluate the effectiveness of our method in traditional big start FSCIL setting, where the base session is considerably large and uses $50\%$ of the classes. We compare the proposed BAMP with four state-of-the-art methods, i.e., FeCAM \cite{goswami2024fecam}, C-FeCAM \cite{goswami2024calibrating}, RanPAC \cite{mcdonnell2024ranpac} and C-RanPAC \cite{goswami2024calibrating}, on six datasets, i.e., CIFAR100 \cite{krizhevsky2009learning}, CUB200 \cite{wah2011caltech}, EuroSAT \cite{helber2019eurosat}, FGVCAircraft \cite{maji2013fine}, Resisc-45 \cite{cheng2017remote} and StanfordCars \cite{krause20133d}. For each dataset, we report the classification accuracy in the final session, denoted as $A_{last}$, alongside the mean incremental accuracy across all sessions, labeled as $A_{inc}$, in Table \ref{tab:results_on_bigstart}. The averages of these metrics over six datasets, termed $mA_{last}$ and $mA_{inc}$, are visualized in Fig. \ref{fig:avg6ds}-A and are also provided in the final two columns of Tab. \ref{tab:results_on_bigstart}. As shown in Fig. \ref{fig:avg6ds}-A and Tab. \ref{tab:results_on_bigstart}, our method BAMP achieves superior performance. In term of average of $A_{last}$ over six datasets, i.e., $mA_{last}$, BAMP achieves $78.09\%$ outperforming four existing approaches by $6.99\%,3.00\%,4.85\%,0.84\%$ respectively. In term of mean $A_{inc}$ over six datasets, i.e., $mA_{inc}$, BAMP achieves $83.84\%$  outperforming four existing approaches by $5.33\%,2.70\%,3.45\%,0.81\%$ respectively. \par
\begin{figure*}[h]  
\centering
\includegraphics[width=\textwidth]{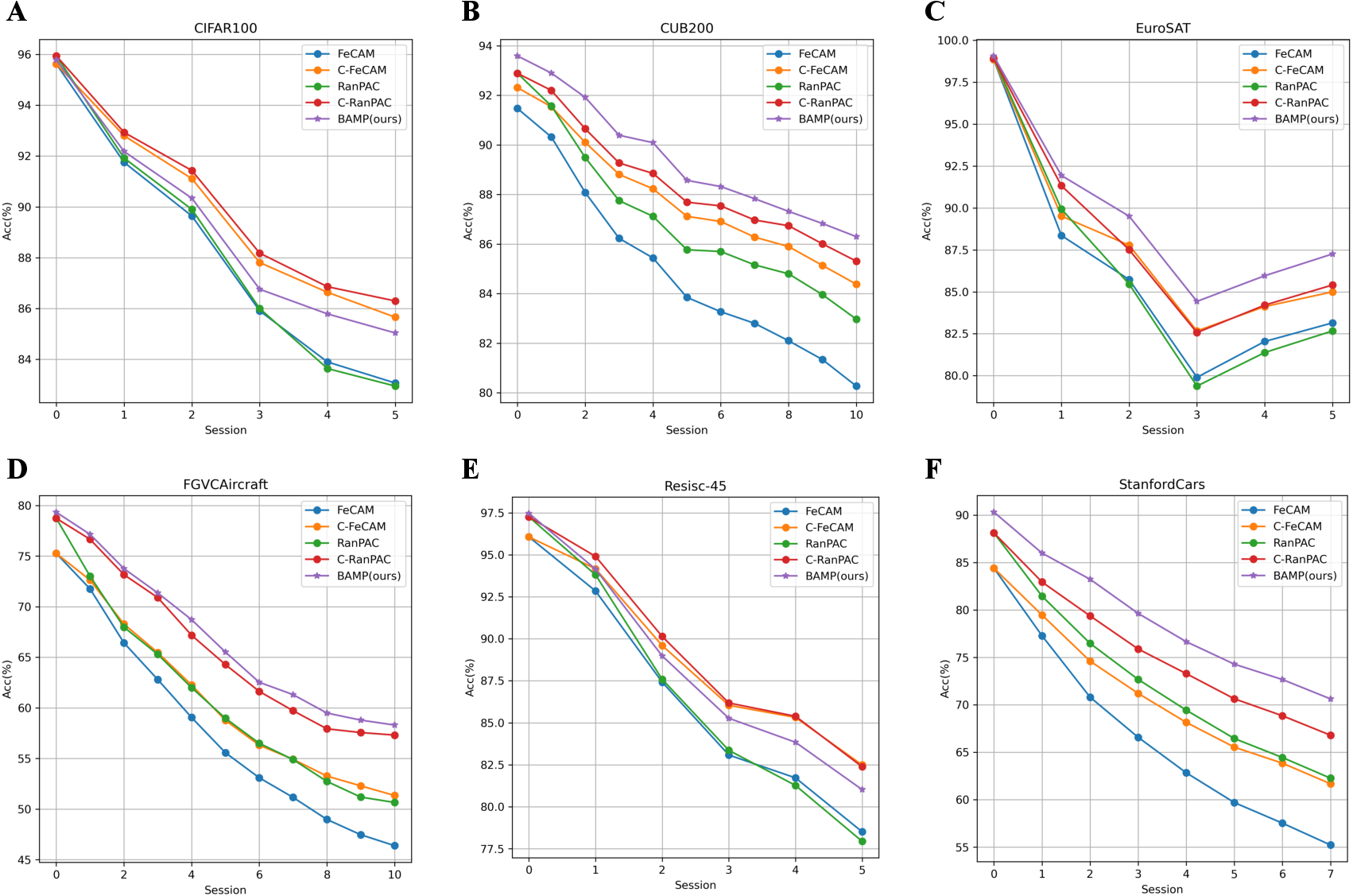}
\caption{Accuracy after each incremental session for big start setting on six datasets.}
\label{fig:resultsOfBigStart}
\end{figure*}

Fig. \ref{fig:resultsOfBigStart} illustrates the classification accuracy after every incremental session under the big start setting across six distinct datasets. The figure evaluates the performance of five different methods. For each dataset, a subplot depicts the accuracy (\%) over successive sessions, highlighting the evolution of accuracy as new data is introduced incrementally. BAMP shows superior performance compared to the other methods on the majority of datasets (four out of six), aligning with the findings presented in Table \ref{tab:results_on_bigstart}. This underscores BAMP's capability to sustain high accuracy throughout the incremental learning process.\par
In sum, we found that the BAMP algorithm outperforms state-of-the-art on traditional big start FSCIL setting.

\begin{table*}[htbp]
\begin{center}
\caption{Performance of BAMP on traditional big start FSCIL setting and its comparison to the baseline methods}\label{tab:results_on_bigstart}
\begin{tabular}{lcccccccccccccc}
\toprule%
\multirow{2}{*}{Method} & \multicolumn{2}{@{}c@{}}{CIFAR100} & \multicolumn{2}{@{}c@{}}{CUB200} & \multicolumn{2}{@{}c@{}}{EuroSAT}& \multicolumn{2}{@{}c@{}}{FGVCAircraft} & \multicolumn{2}{@{}c@{}}{Resisc-45} & \multicolumn{2}{@{}c@{}}{StanfordCars} & \multirow{2}{*}{$mA_{last}$} &\multirow{2}{*}{$mA_{inc}$}\\ 
\cmidrule{2-13}%
 & $A_{last}$ & $A_{inc}$ & $A_{last}$ & $A_{inc}$ & $A_{last}$ & $A_{inc}$ & $A_{last}$ & $A_{inc}$ & $A_{last}$ & $A_{inc}$ &  $A_{last}$ & $A_{inc}$ &  & \\
\midrule
FeCAM \cite{goswami2024fecam}  &83.07 & 88.32 & 80.27 & 85.02   & 83.15 & 86.34 & 46.38 & 57.99 & 78.51 & 86.61& 55.23 & 66.80 & 71.10 & 78.51 \\
C-FeCAM \cite{goswami2024calibrating} & 85.66 & 89.94 & 84.38 & 87.88 & 85.00  & 87.99 & 51.34  & 60.98 &\cellcolor{gray!30}\textbf{82.49} &88.95 & 61.68 & 71.12& 75.09 & 81.14\\
RanPAC \cite{mcdonnell2024ranpac}                  & 82.95 & 88.39 & 82.97 & 87.02 & 82.67 & 86.29 & 50.65 & 61.09 & 77.94 & 86.87 & 62.27 & 72.67 & 73.24 & 80.39 \\
C-RanPAC \cite{goswami2024calibrating} & \cellcolor{gray!30} \textbf{86.30} & \cellcolor{gray!30}\textbf{90.27} & 85.31 & 88.56 & 85.41 & 88.33 & 57.31 & 65.91 & 82.38 & \cellcolor{gray!30}\textbf{89.38} & 66.80 & 75.74 & 77.25 & 83.03 \\
BAMP(ours)    &85.04 	& 89.32 & \cellcolor{gray!30}\textbf{86.30} &\cellcolor{gray!30} \textbf{89.46} &\cellcolor{gray!30} \textbf{87.26} &\cellcolor{gray!30} \textbf{89.70}& \cellcolor{gray!30} \textbf{58.30} & \cellcolor{gray!30} \textbf{66.93} & 81.02 & 88.46 &\cellcolor{gray!30} \textbf{70.63} &\cellcolor{gray!30} \textbf{79.18} & \cellcolor{gray!30} \textbf{78.09} & \cellcolor{gray!30} \textbf{83.84} \\ 
\bottomrule
\end{tabular}
\end{center}
\end{table*}  

\subsection{BAMP Performs Well on Challenging Small Start FSCIL Setting}
\label{subsec:Results_small_start_setting}
To validate the effectiveness of the proposed BAMP on challenging situation, we conduct experiments on small start FSCIL setting, where the dataset is equally split in all tasks and only a few classes are used in base session. The baseline methods and datasets used for comparison are the same as those in the big start setting. For each dataset, classification accuracy after the last session $A_{last}$, and the average of the incremental accuracy from all sessions $A_{inc}$ are reported in Tab. \ref{tab:results_on_smallstart}. Average of $A_{last}$ and $A_{inc}$ over 6 datasets, $mA_{last}$ and $mA_{inc}$ are shown in Fig. \ref{fig:avg6ds}-B, and also reported in the last two columns of Tab. \ref{tab:results_on_smallstart}. It is shown that BAMP achieves better performance. In term of average of $A_{last}$ over six datasets, i.e., $mA_{last}$, BAMP achieves $64.80\%$ outperforming four existing approaches by $9.43\%,6.59\%,4.07\%,1.13\%$ respectively. In term of average of $A_{inc}$ over six datasets, i.e., $mA_{inc}$, BAMP achieves $74.25\%$ outperforming four existing approaches by $8.19\%,5.38\%,4.24\%,1.69\%$ respectively. \par
\begin{figure*}[h]  
\centering
\includegraphics[width=\textwidth]{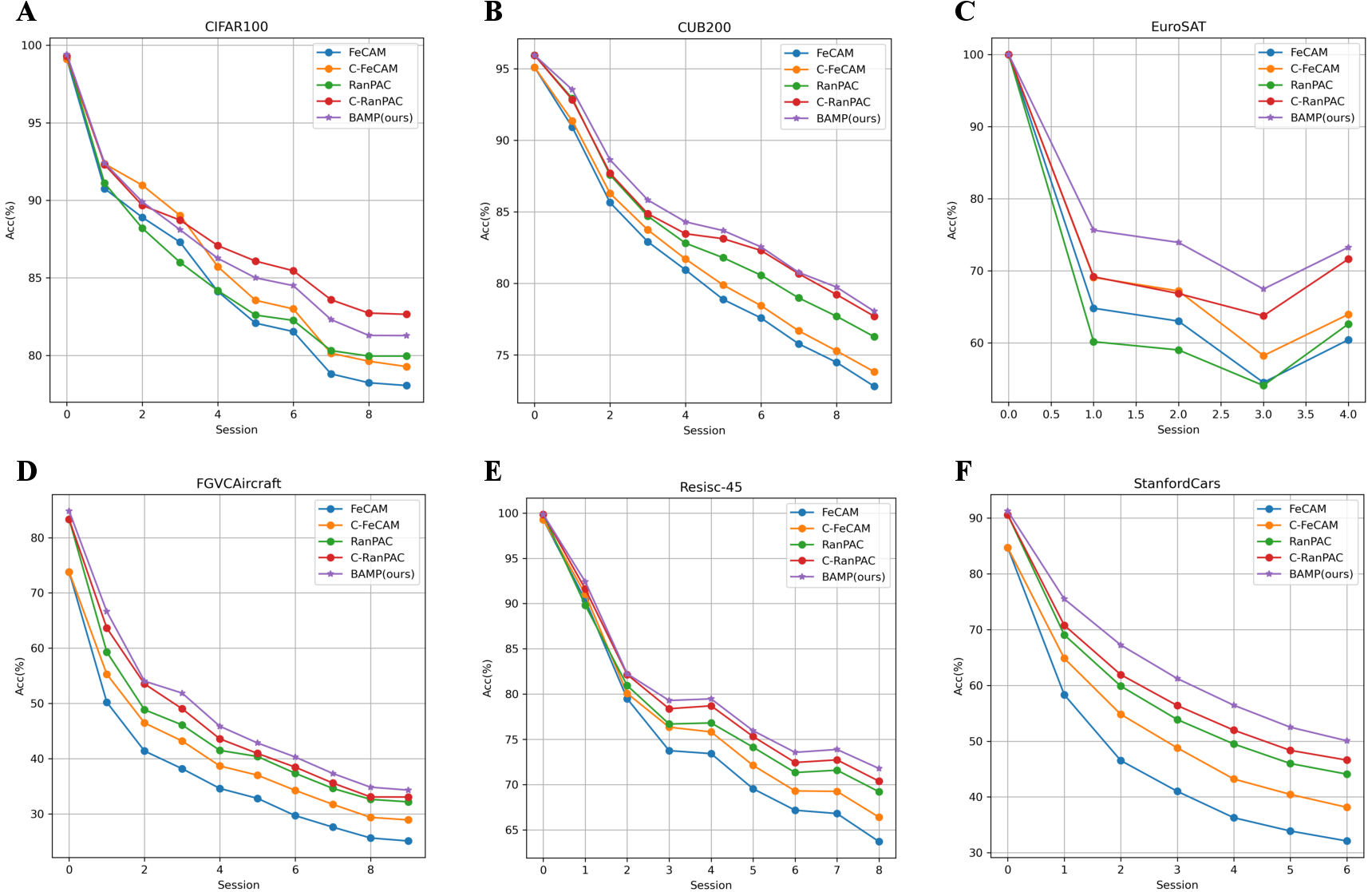}
\caption{Accuracy after each incremental session for small start setting on six datasets.}
\label{fig:resultsOfSmallStart}
\end{figure*}

Fig \ref{fig:resultsOfSmallStart} presents the classification accuracy after each incremental session for the small start setting on six different datasets. The figure compares the performance of five methods. Each subplot shows the accuracy (\%) over the number of sessions on a specific dataset, illustrating how the accuracy changes as new data is incrementally added. Notably, BAMP consistently outperforms the other methods across most of the datasets (five out of six), which is consistent with the presentation in Tab. \ref{tab:results_on_smallstart}, demonstrating its effectiveness in maintaining high accuracy throughout the incremental learning process.

In summary, we found that the BAMP algorithm also performs well on challenging small start FSCIL setting.

\begin{table*}[htbp]
\begin{center}
\caption{Performance of BAMP on challenging small start FSCIL setting and its comparison to the baseline methods}\label{tab:results_on_smallstart}
\begin{tabular}{lcccccccccccccc}
\toprule%
\multirow{2}{*}{Method} & \multicolumn{2}{@{}c@{}}{CIFAR100} & \multicolumn{2}{@{}c@{}}{CUB200} & \multicolumn{2}{@{}c@{}}{EuroSAT}& \multicolumn{2}{@{}c@{}}{FGVCAircraft} & \multicolumn{2}{@{}c@{}}{Resisc-45} & \multicolumn{2}{@{}c@{}}{StanfordCars} & \multirow{2}{*}{$mA_{last}$} & \multirow{2}{*}{$mA_{inc}$}\\ 
\cmidrule{2-13}%
 & $A_{last}$ & $A_{inc}$ & $A_{last}$ & $A_{inc}$ & $A_{last}$ & $A_{inc}$ & $A_{last}$ & $A_{inc}$ & $A_{last}$ & $A_{inc}$ &  $A_{last}$ & $A_{inc}$ &  & \\
\midrule
FeCAM \cite{goswami2024fecam} &78.06 	&84.89 	&72.80 	&81.50 	&60.44 	&68.56 	&25.11 	&37.91 	&63.70 	&75.95 	&32.11 	&47.55 	&55.37 	&66.06 \\
C-FeCAM \cite{goswami2024calibrating} &79.28 	&86.28 	&73.82 	&82.23 	&63.96 	&71.70 	&28.92 	&41.88 	&66.40 	&77.74 	&38.15 	&53.57 	&  58.42 &68.90 \\
RanPAC \cite{mcdonnell2024ranpac}                  &79.96 	&85.39 	&76.27 	&83.93 	&62.63 	&67.19 	&32.19 	&45.63 	&69.21 	&78.93 	&44.09 	&59.00 	&60.73 	&70.01  \\
C-RanPAC \cite{goswami2024calibrating} &\cellcolor{gray!30}\textbf{82.65} 	&\cellcolor{gray!30}\textbf{87.76} 	&77.70 	&84.78 	&71.63 	&74.28 	&33.06 	&47.43 	&70.38 	&80.17 	&46.61 	&60.95 	&63.67 	&72.56  \\
BAMP(ours)    &81.28 	&87.04 	&\cellcolor{gray!30}\textbf{78.05} 	&\cellcolor{gray!30}\textbf{85.30} 	&\cellcolor{gray!30}\textbf{73.26} 	&\cellcolor{gray!30}\textbf{78.06} 	&\cellcolor{gray!30}\textbf{34.32} 	&\cellcolor{gray!30}\textbf{49.28} 	&\cellcolor{gray!30}\textbf{71.78} 	&\cellcolor{gray!30}\textbf{80.94} 	&\cellcolor{gray!30}\textbf{50.08} 	&\cellcolor{gray!30}\textbf{64.90} 	&\cellcolor{gray!30}\textbf{64.80} 	&\cellcolor{gray!30}\textbf{74.25}  \\ 
\bottomrule
\end{tabular}
\end{center}
\end{table*}

\subsection{Ablation Study: All Components of BAMP Works}
\label{subsec:Results_5}

\begin{table*}[htbp]
\begin{center}
\caption{Ablation study. “B” refers to baseline}\label{tab:results_ablation_study}
\begin{tabular}{@{\extracolsep\fill}lccccccc}
\toprule%
\multirow{2}{*}{Method}&\multirow{2}{*}{MoP feature leaning} & \multirow{2}{*}{Calibrated MoP}& \multirow{2}{*}{Soft voting}&\multicolumn{2}{@{}c@{}}{Big start setting} &\multicolumn{2}{@{}c@{}}{Small start setting} \\
\cmidrule{5-6}\cmidrule{7-8}%
 &  & & & $mA_{last}$ & $mA_{inc}$& $mA_{last}$ & $mA_{inc}$\\
\midrule
B1                   &    &    &                         & 71.10&78.51 &55.37	&66.06\\ 
B2                   & \checkmark   &    &               & 71.88&79.29 & 56.21	&67.06\\
B3                   & \checkmark&\checkmark&                         & 75.83 &82.08 & 59.61&70.77\\
B4                   & \checkmark &\checkmark& \checkmark& 78.09 &83.84 & 64.80	&74.25\\
\bottomrule
\end{tabular}
\end{center}
\end{table*}
To understand the contributions of each component in our proposed method, we conducted an ablation study on both big start setting and small start setting. The results are summarized in Tab. \ref{tab:results_ablation_study}, where B refers to the baseline model. We evaluated the performance of our method with different combinations of its key components: MoP feature learning, calibrated MoP, and soft voting. Details are described as follows.
\begin{itemize}
\item Baseline (B1). We take the FeCAM \cite{goswami2024fecam} as the basic model, which is without any additional components. It achieves \(mA_{last}\) of 71.10 and \(mA_{inc}\) of 78.51 in the big start setting, and \(mA_{last}\) of 55.37 and \(mA_{inc}\) of 66.06 in the small start setting.
\item MoP Feature Learning (B2). Adding MoP feature learning improves the performance significantly. In the big start setting, \(mA_{last}\) increases to 71.88 and \(mA_{inc}\) to 79.29. In the small start setting, the improvements are even more pronounced, with \(mA_{last}\) rising to 56.21 and \(mA_{inc}\) to 67.06.
\item MoP Feature Learning \& Calibrated MoP (B3). Incorporating the calibrated MoP further enhances the performance. In the big start setting, \(mA_{last}\) reaches 75.83 and \(mA_{inc}\) reaches 82.08. In the small start setting, \(mA_{last}\) is 59.61 and \(mA_{inc}\) is 70.77.
\item MoP Feature Learning \& Calibrated MoP \& Soft Voting (B4). Finally, adding soft voting to the model leads to the best performance. In the big start setting, \(mA_{last}\) is 78.09 and \(mA_{inc}\) is 83.84. In the small start setting, \(mA_{last}\) is 64.80 and \(mA_{inc}\) is 74.25.
\end{itemize}
These results demonstrate that each component of our method contributes positively to the overall performance, with the combination of all components achieving the highest accuracy in both settings.
\section{Conclusions}
\label{sec:conclusions}
In this study, we introduced Brain-inspired Analogical Mixture Prototypes (BAMP), a novel approach for few-shot class-incremental learning (FSCIL). BAMP combines mixed prototypical feature learning, statistical analogy, and soft voting to effectively alleviate catastrophic forgetting and over fitting while achieving competitive performance in FSCIL scenarios. The mixed prototypical feature learning component learns generalizable feature. The calibration process ensures that the model can adapt its internal representations efficiently when encountering new tasks without compromising previously learned knowledge. The soft voting stage allowing the model to make more informed decisions by integrating multiple sources of information. Experimental results on six benchmark datasets demonstrated that BAMP not only outperforms state-of-the-art on traditional big start FSCIL setting, but also performs well on challenging small start FSCIL setting. 

The success of BAMP opens up multiple avenues for future research. Including: (a) Enhanced Prototypical Learning: Further investigation into how mixed prototypical feature learning can be optimized for larger-scale pre-trained network could lead to even more robust performance. (b) Integration with Other Techniques: Combining BAMP with other continual learning techniques like regularization methods or hybrid models could potentially enhance its effectiveness in handling larger datasets and more complex tasks.

In summary, BAMP represents a significant advancement in addressing the challenges of few-shot class-incremental learning by integrating principles inspired by biological cognition. It contributes to the ongoing research in this field by offering a novel perspective and practical solution for improving few-shot class-incremental learning performance.

\printbibliography 

@article{lu2024learning_iclr,
  title={Learning with mixture of prototypes for out-of-distribution detection},
  author={Lu, Haodong and Gong, Dong and Wang, Shuo and Xue, Jason and Yao, Lina and Moore, Kristen},
  journal={arXiv preprint arXiv:2402.02653},
  year={2024}
}

@Article{ijcv_2024_Revisiting,
  author   = {Zhou, Da-Wei and Cai, Zi-Wen and Ye, Han-Jia and Zhan, De-Chuan and Liu, Ziwei},
  title    = {Revisiting Class-Incremental Learning with Pre-Trained Models: Generalizability and Adaptivity are All You Need},
  journal  = {International Journal of Computer Vision},
  year     = {2024}, 
  issn     = {1573-1405},
  refid    = {Zhou2024},
  url      = {https://doi.org/10.1007/s11263-024-02218-0},
}

@article{wang2024few,
  title={Few-shot class-incremental learning via training-free prototype calibration},
  author={Wang, Qi-Wei and Zhou, Da-Wei and Zhang, Yi-Kai and Zhan, De-Chuan and Ye, Han-Jia},
  journal={Advances in Neural Information Processing Systems},
  volume={36},
  year={2024}
}

@inproceedings{goswami2024calibrating,
  title={Calibrating Higher-Order Statistics for Few-Shot Class-Incremental Learning with Pre-trained Vision Transformers},
  author={Goswami, Dipam and Twardowski, Bart{\l}omiej and Van De Weijer, Joost},
  booktitle={Proceedings of the IEEE/CVF Conference on Computer Vision and Pattern Recognition},
  pages={4075--4084},
  year={2024}
}

@article{goswami2024fecam,
  title={Fecam: Exploiting the heterogeneity of class distributions in exemplar-free continual learning},
  author={Goswami, Dipam and Liu, Yuyang and Twardowski, Bart{\l}omiej and van de Weijer, Joost},
  journal={Advances in Neural Information Processing Systems},
  volume={36},
  year={2024}
}

@article{mcdonnell2024ranpac,
  title={Ranpac: Random projections and pre-trained models for continual learning},
  author={McDonnell, Mark D and Gong, Dong and Parvaneh, Amin and Abbasnejad, Ehsan and van den Hengel, Anton},
  journal={Advances in Neural Information Processing Systems},
  volume={36},
  year={2024}
}

@inproceedings{opfer2005varying,
  title={A Varying Abstraction Model for Categorization},
  author={Opfer, John E and Siegler, Robert S},
  booktitle={Proceedings of the Annual Meeting of the Cognitive Science Society},
  volume={27},
  number={27},
  year={2005}
}

@article{gentner2017analogy,
  title={Analogy and abstraction},
  author={Gentner, Dedre and Hoyos, Christian},
  journal={Topics in cognitive science},
  volume={9},
  number={3},
  pages={672--693},
  year={2017},
  publisher={Wiley Online Library}
}

@mastersthesis{krizhevsky2009learning,
    author = {Krizhevsky, Alex and Hinton, Geoffrey and others},
    title = {Learning multiple layers of features from tiny images},
    school = {Univ. Toronto},
    year = {2009}
}

@misc{wah2011caltech,
  author		= "Wah, Catherine and Branson, Steve and Welinder, Peter and Perona, Pietro and Belongie, Serge",
  title			= "The caltech-ucsd birds-200-2011 dataset", 
  year			= "2011",
  note			= "Available: \url{http://www.vision.caltech.edu/visipedia/CUB-200.html}"
}

@article{maji2013fine,
  title={Fine-grained visual classification of aircraft},
  author={Maji, Subhransu and Rahtu, Esa and Kannala, Juho and Blaschko, Matthew and Vedaldi, Andrea},
  journal={arXiv preprint arXiv:1306.5151},
  year={2013}
}

@inproceedings{krause20133d,
  title={3d object representations for fine-grained categorization},
  author={Krause, Jonathan and Stark, Michael and Deng, Jia and Fei-Fei, Li},
  booktitle={Proceedings of the IEEE international conference on computer vision workshops},
  pages={554--561},
  year={2013}
}

@article{helber2019eurosat,
  title={Eurosat: A novel dataset and deep learning benchmark for land use and land cover classification},
  author={Helber, Patrick and Bischke, Benjamin and Dengel, Andreas and Borth, Damian},
  journal={IEEE Journal of Selected Topics in Applied Earth Observations and Remote Sensing},
  volume={12},
  number={7},
  pages={2217--2226},
  year={2019},
  publisher={IEEE}
}

@article{cheng2017remote,
  title={Remote sensing image scene classification: Benchmark and state of the art},
  author={Cheng, Gong and Han, Junwei and Lu, Xiaoqiang},
  journal={Proceedings of the IEEE},
  volume={105},
  number={10},
  pages={1865--1883},
  year={2017},
  publisher={IEEE}
}

@conference{dosovitskiy2020image,
    author = {Dosovitskiy, Alexey},
    booktitle = {ICLR},
    title = {An image is worth 16x16 words: Transformers for image recognition at scale},
    year = {2020}
}

@article{chen2022adaptformer,
  title={Adaptformer: Adapting vision transformers for scalable visual recognition},
  author={Chen, Shoufa and Ge, Chongjian and Tong, Zhan and Wang, Jiangliu and Song, Yibing and Wang, Jue and Luo, Ping},
  journal={Advances in Neural Information Processing Systems},
  volume={35},
  pages={16664--16678},
  year={2022}
}

@book{mardia2000directional,
  title={Directional Statistics},
  author={Mardia, Kanti V. and Jupp, Peter E.},
  volume={2},
  year={2000},
  publisher={Wiley Online Library}
}

@inproceedings{wang2020understanding,
  title={Understanding contrastive representation learning through alignment and uniformity on the hypersphere},
  author={Wang, Tongzhou and Isola, Phillip},
  booktitle={International Conference on Machine Learning},
  pages={9929--9939},
  year={2020},
  organization={PMLR}
}

@misc{timm_library,
  author       = {Ross Wightman},
  title        = {{Timm: PyTorch Image Models}},
  year         = {2020},
  publisher    = {GitHub},
  journal      = {GitHub repository},
  howpublished = {\url{https://github.com/rwightman/pytorch-image-models}}  
}

@inproceedings{qiu2023semantic,
  title={Semantic-visual guided transformer for few-shot class-incremental learning},
  author={Qiu, Wenhao and Fu, Sichao and Zhang, Jingyi and Lei, Chengxiang and Peng, Qinmu},
  booktitle={2023 IEEE International Conference on Multimedia and Expo (ICME)},
  pages={2885--2890},
  year={2023},
  organization={IEEE}
}

@inproceedings{park2024pre,
  title={Pre-trained Vision and Language Transformers Are Few-Shot Incremental Learners},
  author={Park, Keon-Hee and Song, Kyungwoo and Park, Gyeong-Moon},
  booktitle={Proceedings of the IEEE/CVF Conference on Computer Vision and Pattern Recognition},
  pages={23881--23890},
  year={2024}
}

@article{ran2024brain,
  title={Brain-inspired fast-and slow-update prompt tuning for few-shot class-incremental learning},
  author={Ran, Hang and Gao, Xingyu and Li, Lusi and Li, Weijun and Tian, Songsong and Wang, Gang and Shi, Hailong and Ning, Xin},
  journal={IEEE Transactions on Neural Networks and Learning Systems},
  year={2024},
  publisher={IEEE}
}

@article{wang2021few,
  title={Few-shot Continual Learning: a Brain-inspired Approach},
  author={Wang, Liyuan and Li, Qian and Zhong, Yi and Zhu, Jun},
  journal={arXiv preprint arXiv:2104.09034},
  year={2021}
}

@article{zhao2021mgsvf,
  title={Mgsvf: Multi-grained slow versus fast framework for few-shot class-incremental learning},
  author={Zhao, Hanbin and Fu, Yongjian and Kang, Mintong and Tian, Qi and Wu, Fei and Li, Xi},
  journal={IEEE Transactions on Pattern Analysis and Machine Intelligence},
  volume={46},
  number={3},
  pages={1576--1588},
  year={2021},
  publisher={IEEE}
}

@article{tian2024survey,
  title={A survey on few-shot class-incremental learning},
  author={Tian, Songsong and Li, Lusi and Li, Weijun and Ran, Hang and Ning, Xin and Tiwari, Prayag},
  journal={Neural Networks},
  volume={169},
  pages={307--324},
  year={2024},
  publisher={Elsevier}
}

@article{zhang2025few,
  title={Few-Shot Class-Incremental Learning for Classification and Object Detection: A Survey},
  author={Zhang, Jinghua and Liu, Li and Silv{\'e}n, Olli and Pietik{\"a}inen, Matti and Hu, Dewen},
  journal={IEEE Transactions on Pattern Analysis and Machine Intelligence},
  year={2025},
  publisher={IEEE}
}

@inproceedings{kukleva2021generalized,
  title={Generalized and incremental few-shot learning by explicit learning and calibration without forgetting},
  author={Kukleva, Anna and Kuehne, Hilde and Schiele, Bernt},
  booktitle={Proceedings of the IEEE/CVF international conference on computer vision},
  pages={9020--9029},
  year={2021}
}

@inproceedings{zhu2022feature,
  title={Feature distribution distillation-based few shot class incremental learning},
  author={Zhu, Juntao and Yao, Guangle and Zhou, Wenlong and Zhang, Guiyu and Ping, Wang and Zhang, Wei},
  booktitle={2022 5th International Conference on Pattern Recognition and Artificial Intelligence (PRAI)},
  pages={108--113},
  year={2022},
  organization={IEEE}
}

@inproceedings{liu2022few,
  title={Few-shot class-incremental learning via entropy-regularized data-free replay},
  author={Liu, Huan and Gu, Li and Chi, Zhixiang and Wang, Yang and Yu, Yuanhao and Chen, Jun and Tang, Jin},
  booktitle={European Conference on Computer Vision},
  pages={146--162},
  year={2022},
  organization={Springer}
}

@inproceedings{shankarampeta2021few,
  title={Few-Shot Class Incremental Learning with Generative Feature Replay.},
  author={Shankarampeta, Abhilash Reddy and Yamauchi, Koichiro},
  booktitle={ICPRAM},
  pages={259--267},
  year={2021}
}

@inproceedings{zhou2022forward,
  title={Forward compatible few-shot class-incremental learning},
  author={Zhou, Da-Wei and Wang, Fu-Yun and Ye, Han-Jia and Ma, Liang and Pu, Shiliang and Zhan, De-Chuan},
  booktitle={Proceedings of the IEEE/CVF conference on computer vision and pattern recognition},
  pages={9046--9056},
  year={2022}
}

@inproceedings{peng2022few,
  title={Few-shot class-incremental learning from an open-set perspective},
  author={Peng, Can and Zhao, Kun and Wang, Tianren and Li, Meng and Lovell, Brian C},
  booktitle={European Conference on Computer Vision},
  pages={382--397},
  year={2022},
  organization={Springer}
}

@inproceedings{zhang2021few,
  title={Few-shot incremental learning with continually evolved classifiers},
  author={Zhang, Chi and Song, Nan and Lin, Guosheng and Zheng, Yun and Pan, Pan and Xu, Yinghui},
  booktitle={Proceedings of the IEEE/CVF conference on computer vision and pattern recognition},
  pages={12455--12464},
  year={2021}
}

@inproceedings{zhu2021self,
  title={Self-promoted prototype refinement for few-shot class-incremental learning},
  author={Zhu, Kai and Cao, Yang and Zhai, Wei and Cheng, Jie and Zha, Zheng-Jun},
  booktitle={Proceedings of the IEEE/CVF conference on computer vision and pattern recognition},
  pages={6801--6810},
  year={2021}
}

@inproceedings{tao2020few,
  title={Few-shot class-incremental learning},
  author={Tao, Xiaoyu and Hong, Xiaopeng and Chang, Xinyuan and Dong, Songlin and Wei, Xing and Gong, Yihong},
  booktitle={Proceedings of the IEEE/CVF conference on computer vision and pattern recognition},
  pages={12183--12192},
  year={2020}
}

@article{yang2022dynamic,
  title={Dynamic support network for few-shot class incremental learning},
  author={Yang, Boyu and Lin, Mingbao and Zhang, Yunxiao and Liu, Binghao and Liang, Xiaodan and Ji, Rongrong and Ye, Qixiang},
  journal={IEEE Transactions on Pattern Analysis and Machine Intelligence},
  volume={45},
  number={3},
  pages={2945--2951},
  year={2022},
  publisher={IEEE}
}

@inproceedings{zhao2023few,
  title={Few-shot class-incremental learning via class-aware bilateral distillation},
  author={Zhao, Linglan and Lu, Jing and Xu, Yunlu and Cheng, Zhanzhan and Guo, Dashan and Niu, Yi and Fang, Xiangzhong},
  booktitle={Proceedings of the IEEE/CVF conference on computer vision and pattern recognition},
  pages={11838--11847},
  year={2023}
}

@article{zhou2022few,
  title={Few-shot class-incremental learning by sampling multi-phase tasks},
  author={Zhou, Da-Wei and Ye, Han-Jia and Ma, Liang and Xie, Di and Pu, Shiliang and Zhan, De-Chuan},
  journal={IEEE Transactions on Pattern Analysis and Machine Intelligence},
  volume={45},
  number={11},
  pages={12816--12831},
  year={2022},
  publisher={IEEE}
}

@inproceedings{mazumder2021few,
  title={Few-shot lifelong learning},
  author={Mazumder, Pratik and Singh, Pravendra and Rai, Piyush},
  booktitle={Proceedings of the AAAI Conference on Artificial Intelligence},
  volume={35},
  number={3},
  pages={2337--2345},
  year={2021}
}

@inproceedings{hersche2022constrained,
  title={Constrained few-shot class-incremental learning},
  author={Hersche, Michael and Karunaratne, Geethan and Cherubini, Giovanni and Benini, Luca and Sebastian, Abu and Rahimi, Abbas},
  booktitle={Proceedings of the IEEE/CVF conference on computer vision and pattern recognition},
  pages={9057--9067},
  year={2022}
}

@inproceedings{zheng2021few,
  title={Few-shot class-incremental learning with meta-learned class structures},
  author={Zheng, Guangtao and Zhang, Aidong},
  booktitle={2021 International Conference on Data Mining Workshops (ICDMW)},
  pages={421--430},
  year={2021},
  organization={IEEE}
}

@inproceedings{dong2021few,
  title={Few-shot class-incremental learning via relation knowledge distillation},
  author={Dong, Songlin and Hong, Xiaopeng and Tao, Xiaoyu and Chang, Xinyuan and Wei, Xing and Gong, Yihong},
  booktitle={Proceedings of the AAAI Conference on Artificial Intelligence},
  volume={35},
  number={2},
  pages={1255--1263},
  year={2021}
}

@inproceedings{cheraghian2021semantic,
  title={Semantic-aware knowledge distillation for few-shot class-incremental learning},
  author={Cheraghian, Ali and Rahman, Shafin and Fang, Pengfei and Roy, Soumava Kumar and Petersson, Lars and Harandi, Mehrtash},
  booktitle={Proceedings of the IEEE/CVF conference on computer vision and pattern recognition},
  pages={2534--2543},
  year={2021}
}
\end{document}